\documentclass[11pt]{article}

\usepackage[final]{acl}

\usepackage{times}
\usepackage{latexsym}

\usepackage[T1]{fontenc}
\usepackage[utf8]{inputenc}
\usepackage{array}

\usepackage{microtype}

\usepackage{inconsolata}

\usepackage{graphicx}

\usepackage{amsmath}
\usepackage{arydshln}

\usepackage{multirow}
\usepackage{booktabs}
\usepackage{siunitx}
\usepackage{float}
\usepackage{placeins}
\usepackage{dblfloatfix}
\usepackage{adjustbox}
\usepackage{makecell} 
\usepackage{tablefootnote}
\usepackage{pdfpages}

\usepackage{tikz}
\newcommand*\circled[1]{\tikz[baseline=(char.base)]{
            \node[shape=circle,draw,inner sep=2pt] (char) {#1};}}
            
\title{AILS-NTUA at SemEval-2026 Task 3: Efficient Dimensional Aspect-Based Sentiment Analysis}

\author{Stavros Gazetas \hspace{0.8em} Giorgos Filandrianos  \hspace{0.8em} Maria Lymperaiou \\
\textbf{Paraskevi Tzouveli \quad Athanasios Voulodimos \quad Giorgos Stamou} \\
School of Electrical and Computer Engineering, AILS Laboratory \\
National Technical University of Athens \\
\href{mailto:stavrosgazetas@hotmail.com}{\texttt{stavrosgaz9@gmail.com}}, \href{mailto:geofila@ails.ece.ntua.gr}{\texttt{\{geofila,}} \href{mailto:marialymp@ails.ece.ntua.gr}{\texttt{marialymp\}@ails.ece.ntua.gr}} \\
\href{mailto:tpar@mail.ntua.gr}{\texttt{tpar@mail.ntua.gr}},
\href{mailto:thanosv@mail.ntua.gr}{\texttt{thanosv@mail.ntua.gr}}, \href{mailto:gstam@cs.ntua.gr}{\texttt{gstam@cs.ntua.gr}}}

\begin{document}
\maketitle
\begin{abstract}
% In this paper, we present our submissions for Track-A of SemEval-2026 Task 3: Dimensional Aspect Based Sentiment Analysis (DimABSA). Participants were requested to tackle Dimensional Aspect Sentiment Regression (DimASR), Dimensional Aspect Sentiment Triplet Extraction (DimASTE), and Dimensional Aspect Sentiment Quad Prediction (DimASQP) in a multilingual, multi-domain setting. Our approach included fine-tuning language-appropriate encoder backbones for DimASR and language-specific instruction fine-tuning of LLMs using LoRA for DimASTE and DimASQP. Overall, our parameter-efficient models reduce training and inference costs while achieving competitive results and outperforming the provided baselines in most settings\footnote{Code: \href{https://github.com/stavgaz/Semeval2026-Efficient-DimABSA}{stavgaz/Semeval2026-Efficient-DimABSA}}.

In this paper, we present AILS-NTUA system for Track-A of SemEval-2026 Task 3 on Dimensional Aspect-Based Sentiment Analysis (DimABSA), which encompasses three complementary problems: Dimensional Aspect Sentiment Regression (DimASR), Dimensional Aspect Sentiment Triplet Extraction (DimASTE), and Dimensional Aspect Sentiment Quadruplet Prediction (DimASQP) within a multilingual and multi-domain framework. Our methodology combines fine-tuning of language-appropriate encoder backbones for continuous aspect-level sentiment prediction with language-specific instruction tuning of large language models using LoRA for structured triplet and quadruplet extraction. This unified yet task-adaptive design emphasizes parameter-efficient specialization across languages and domains, enabling reduced training and inference requirements while maintaining strong effectiveness. Empirical results demonstrate that the proposed models achieve competitive performance and consistently surpass the provided baselines across most evaluation settings\footnote{Code: \href{https://github.com/stavgaz/Semeval2026-Efficient-DimABSA}{stavgaz/Semeval2026-Efficient-DimABSA}}.

% The code is available on GitHub \footnote{\href{https://github.com/stavgaz/Semeval2026-Efficient-DimABSA}{stavgaz/Semeval2026-Efficient-DimABSA}}.
\end{abstract}

\section{Introduction}
Aspect-Based Sentiment Analysis (ABSA) seeks to discern the sentiment directed at a specific aspect of an entity, facilitating a fine-grained analysis that transcends sentiment polarity. In recent years, research has increasingly moved from predicting categorical sentiment for isolated elements toward structured ABSA formulations that extract richer opinion structures, such as triplets in Aspect Sentiment Triplet Extraction (ASTE) \cite{Peng_Xu_Bing_Huang_Lu_Si_2020} and quadruplets in Aspect Sentiment Quad Prediction (ASQP) \cite{zhang-etal-2021-aspect-sentiment}, built around aspect terms/categories and opinion expressions with sentiment polarity \cite{zhang2022surveyaspectbasedsentimentanalysis}. By replacing the sentiment polarity with continuous valence–arousal (VA) scores, where valence reflects positivity/negativity and arousal reflects activation/intensity \cite{VA_rus}, more nuanced emotional information can be provided, as explored in a shared task \cite{lee-etal-2024-overview-sighan}.

SemEval-2026 Task 3: Dimensional Aspect-Based Sentiment Analysis (DimABSA) extends ABSA to a multilingual and multi-domain setting by utilizing VA annotations to facilitate analysis at both the aspect and sentiment levels \cite{yu-etal-2026-semeval}. Our work revolves around Track-A \cite{lee2026dimabsabuildingmultilingualmultidomain}, which includes three subtasks: Subtask 1 - Dimensional Aspect Sentiment Regression (DimASR), Subtask 2 - Dimensional Aspect Sentiment Triplet Extraction (DimASTE), and Subtask 3 - Dimensional Aspect Sentiment Quad Prediction (DimASQP), which we explore in this work.

% To tackle these subtasks, we adopt two complementary strategies. DimASR is treated as an aspect-conditioned regression problem, while DimASTE and DimASQP are treated as structured extraction tasks. Our system emphasizes computational efficiency via lightweight, parameter-efficient models, aiming to reduce training and inference costs while preserving performance. Based on this, we make the following contributions:

% \begin{itemize}
%     \item For DimASR, we fine-tune pretrained transformer encoders separately for each language/domain dataset to predict continuous VA scores.
%     \item For DimASTE and DimASQP, we perform per-language and per-domain instruction fine-tuning using LoRA on Llama and Qwen family models, obtaining competitive performance with backbones of $\leq$14B parameters.
%     \item We additionally conduct a study on the translation of low-resource languages into English for training, discovering useful insights. 
% \end{itemize}

To address the three subtasks, we propose a unified yet task-specialized framework that combines aspect-conditioned regression with instruction-tuned structured generation: DimASR is modeled as continuous VA regression, while DimASTE and DimASQP are cast as constrained JSON generation. Our contributions are:
\circled{1} A parameter-efficient multilingual regression framework for DimASR, fine-tuning language-appropriate encoders per language--domain pair to predict valence--arousal scores, outperforming the provided baselines in most settings.
\circled{2} A unified LoRA-based instruction-tuning pipeline for DimASTE and DimASQP on Llama and Qwen models ($\leq$14B), evaluated under zero-, few-, and supervised setups, achieving competitive or superior cF1 compared to larger fully fine-tuned LLMs.
\circled{3} An empirical study of translation-based cross-lingual transfer, analyzing the trade-offs between multilingual adaptation and translation-induced noise in lower-resource settings.

% The code is available on GitHub \footnote{\href{https://github.com/stavgaz/Semeval2026-Efficient-DimABSA}{stavgaz/Semeval2026-Efficient-DimABSA}}.

\section{Related Work}

\textbf{Tagging and span-based structured ABSA.}
Earlier work on structured ABSA commonly revolved around using sequence labeling, tagging, and span-based methods to jointly extract triplets (aspect, opinion, sentiment) and quadruplets (aspect, category, opinion, sentiment). Regarding ASTE, approaches such as position-aware tagging \cite{xu-etal-2020-position}, grid-based token-pair labeling \cite{wu-etal-2020-grid}, and span-based formulations \cite{xu-etal-2021-learning} were implemented. As for ASQP, non-generative approaches frequently employ table-filling or grid-tagging to predict relations among the elements in a unified manner \cite{zhou2023unifiedonestepsolutionaspect}, building on earlier quadruple settings such as ACOS \cite{cai-etal-2021-aspect}. These strategies motivate further research that substitutes discrete labeling methods with serialized tuple generation.

\textbf{Generative structured ABSA.}
Structured ABSA has increasingly been formulated as text generation, where models decode serialized sentiment structures directly from the input text. Early work used encoder-decoder models such as T5/BART to generate the targets in unified text-to-text frameworks \cite{zhang-etal-2021-towards-generative, yan-etal-2021-unified}. 
More recent work has utilized large language models (LLMs) for ABSA under zero-shot and few-shot prompting, as well as instruction-style fine-tuning, showing promise across different domains and languages \cite{Wu_2025, wang-etal-2024-context, zhou2024comprehensiveevaluationlargelanguage, simmering2023largelanguagemodelsaspectbased}. This groundwork motivated our use of instruction-tuned LLMs for structured ABSA in the dimensional (VA) setting of DimABSA.

\textbf{Dimensional ABSA}
The VA annotations have recently been adopted in ABSA as a way to extract fine-grained information. Most notably, the SIGHAN-2024 dimABSA shared task studied dimensional ABSA with subtasks covering intensity prediction and structured triplet/quadruplet extraction in VA space \cite{lee-etal-2024-overview-sighan}. In this shared task, many approaches were proposed for the Chinese language, including span-based extraction with contrastive learning \cite{tong-wei-2024-cciiplab}, paraphrase/generation-style prediction \cite{jiang-etal-2024-jn}, and LLM-based approaches, including few-shot in-context learning and fine-tuning \cite{meng-etal-2024-ds,xu-etal-2024-hitsz}.

\section{Task and Dataset Description}

\subsection{Task Definition}

% The task is defined over the following elements: Aspect Term ($A$), Aspect Category ($C$, formatted as \texttt{ENTITY\#ATTRIBUTE}), Opinion Term ($O$), and Valence-Arousal ($VA$, formatted as \texttt{$V$\#$A$}, where $V$, $A$ $\in [1.00, 9.00]$, and outputs are rounded to two decimals). Task format and examples are provided in App.~\ref{app:format}.

The task is defined over the following elements: Aspect Term ($A$), Aspect Category ($C$), Opinion Term ($O$), and Valence–Arousal ($VA$). The \textit{Aspect Category} follows a hierarchical schema of the form \texttt{ENTITY\#ATTRIBUTE}, where \textit{ENTITY} denotes the general aspect entity (e.g., \textit{FOOD}, \textit{SERVICE}) and \textit{ATTRIBUTE} specifies the evaluated property of that entity (e.g., \textit{QUALITY}, \textit{PRICE}), with the symbol \texttt{\#} acting as a delimiter between the two semantic levels. The Valence–Arousal pair is represented as \texttt{V\#A}, where $V, A \in [1.00, 9.00]$ and values are reported with two-decimal precision. Task format and examples are provided in App.~\ref{app:format}.

Track-A\footnote{Official Task repository: \href{https://github.com/DimABSA/DimABSA2026}{DimABSA/DimABSA2026}} includes the three previously mentioned subtasks: DimASR: given a text and one or more target aspects, predict a VA score for each aspect. DimASTE: given a text, extract all triplets $(A,O,VA)$. DimASQP: given a text, extract all quadruplets $(A,C,O,VA)$. All textual outputs are case-sensitive.

\subsection{Dataset Description}
% The DimABSA benchmark contains six languages: \textit{Chinese(ZHO), English (ENG), Japanese (JPN), Russian (RUS), Tatar (TAT) and Ukrainian (UKR)} and four domains: \textit{Restaurant, Laptop, Hotel and Finance}. All existing language/domain combinations have Train/Dev/Test splits. Each dataset instance is a text review for the specific domain. Finance datasets are provided only for DimASR. Some datasets showcased \emph{NULL} labels (implicit sentiment) in the Train set, which affected training. There were also shifts in review lengths, in the number of quads per review, and in category representation among the splits. We provide dataset statistics and exploratory data analysis (EDA) in App.~\ref{app:eda}.

The DimABSA benchmark comprises six languages—\textit{Chinese (ZHO), English (ENG), Japanese (JPN), Russian (RUS), Tatar (TAT), and Ukrainian (UKR)}—and spans four application domains: \textit{Restaurant, Laptop, Hotel, and Finance}. For each available language–domain configuration, the dataset is organized into dedicated \textit{training}, \textit{validation}, and \textit{test} partitions. Individual instances correspond to domain-specific textual reviews annotated with aspect-level dimensional sentiment information. Notably, the Finance domain is available exclusively for the DimASR subtask.

Certain training partitions—namely the English Restaurant and Laptop datasets, the Chinese Restaurant dataset, and the Japanese Hotel dataset—contain \emph{NULL} annotations corresponding to implicitly expressed sentiment targets, resulting in increased difficulty during model training. Furthermore, we observe distributional variability across dataset partitions, including differences in review length, sentiment structure density (i.e., quadruplets per review), and category frequency. Detailed dataset statistics and exploratory data analysis are provided in App.~\ref{app:eda}.

\subsection{Evaluation Metrics}
\paragraph{DimASR} is evaluated using Root Mean Squared Error (RMSE) between the predicted and gold valence-arousal (VA) scores, computed over both dimensions. 

\paragraph{DimASTE and DimASQP} are evaluated using continuous F1 (cF1), which unifies the exact-match evaluation of categorical elements with the accuracy of VA prediction. A prediction contributes only if its categorical elements exactly match a gold annotation in the same sentence (i.e., $(A,O)$ for triplets and $(A,C,O)$ for quadruplets).

\section{System Overview}

\subsection{Aspect-Based Regression for DimASR}
For DimASR, we fine-tune pretrained transformer encoders for aspect-conditioned regression. Given an input sentence $x$ and a target aspect $a$, we concatenate them into a single sequence (e.g., \texttt{Aspect: $a$. Sentence: $x$.}) and feed it to the encoder backbone. We obtain a pooled representation and predict valence and arousal using two scalar regression heads. To train the models, we optimize a weighted combination of Mean Squared Error (MSE) and Concordance Correlation Coefficient (CCC) \cite{lin1989ccc}, and we add a VA-guided triplet regularizer using the standard hinge triplet objective \cite{schroff2015facenet}. We use one backbone per language and fine-tune separate models for each domain within that language. All the datasets are flattened to contain entries in the form of Review/Aspect. At inference, each review's VA scores are grouped back. A more thorough explanation is provided in App.~\ref{app:reg}.

\paragraph{Language backbones.}
% We use language-appropriate transformer encoders from the BERT, RoBERTa, and DeBERTa families \cite{devlin2019bert,liu2019roberta,he2020deberta} and employ XLM-R \cite{conneau2020xlmr} for languages without a dedicated backbone. Specifically, we adopt DeBERTa as the backbone for English\footnote{\href{https://huggingface.co/yangheng/deberta-v3-base-absa-v1.1}{yangheng/deberta-v3-base-absa-v1.1}} and Japanese\footnote{\href{https://huggingface.co/ku-nlp/deberta-v3-base-japanese}{ku-nlp/deberta-v3-base-japanese}}, RoBERTa for Chinese\footnote{\href{https://huggingface.co/hfl/chinese-roberta-wwm-ext}{hfl/chinese-roberta-wwm-ext}}, a BERT-style model for Russian\footnote{\href{https://huggingface.co/DeepPavlov/rubert-base-cased}{DeepPavlov/rubert-base-cased}}, and XLM-R for Ukrainian and Tatar\footnote{\href{https://huggingface.co/FacebookAI/xlm-roberta-base}{FacebookAI/xlm-roberta-base}}.
We use language-appropriate transformer encoders from the BERT, RoBERTa, and DeBERTa families \cite{devlin2019bert,liu2019roberta,he2020deberta} and employ XLM-R \cite{conneau2020xlmr} for languages without a dedicated backbone. Specifically, we use DeBERTa as the backbone for English and Japanese, RoBERTa for Chinese, a BERT-style model for Russian, and XLM-R for Ukrainian and Tatar. Details about the selected backbones are provided in App.~\ref{app:reg}.

\subsection{Instruction-Tuned LLM Generation for DimASTE and DimASQP}
In order to extract the sentiment elements from the review, we approach DimASTE and DimASQP as constrained text generation. Given an input review, the model generates a JSON-formatted list of sentiment structures. This enables joint prediction of all required elements in a single decoding pass and avoids multi-stage extraction pipelines.

\paragraph{Unified prompting with and without categories.}
Both subtasks share the same instruction-tuning pipeline and differ only in whether aspect categories are provided and predicted. For DimASQP, the prompt includes the domain-specific list of valid categories, whereas for DimASTE the category list is omitted. For partitions containing \emph{NULL} labels, we design language-specific instructions that discourage the model from predicting \emph{NULL} as an easy solution when the aspect and/or opinion is implicit in the review. All instructions are written in the same language as the input data. The exact templates used for training and inference are reported in App.~\ref{app:prompts}.

% Both subtasks share the same instruction-tuning pipeline and differ only in whether aspect categories are provided and predicted: DimASQP prompts include the domain-specific category list, whereas DimASTE prompts omit it. For partitions containing \emph{NULL} labels, we design language-specific instructions that discourage the model from predicting \emph{NULL} when the aspect and/or opinion is implicit in the review. All instructions are written in the same language as the input data, and the exact templates used for training and inference are reported in App.~\ref{app:prompts}.

\paragraph{Structured outputs and VA handling.}
To reduce formatting errors and simplify post-processing, we split $VA$ into explicit numeric keys (Valence, Arousal) in the generated JSON during training. Gold labels are converted from the original \texttt{$V$\#$A$} strings into floats and serialized with two-decimal formatting. At inference time, we parse the generated JSON, constrain $VA$ values to the valid $[1.00,9.00]$ range, and then map predictions back to the required submission format.

% \begin{figure}[t!]
%     \centering
%     \includegraphics[width=0.5\textwidth]{images/system_fig.png}
%     \caption{DimASTE and DimASQP system pipeline.}
%     \label{fig:MODEL}
% \end{figure}

\paragraph{Backbones and parameter-efficient fine-tuning.}
We instruction fine-tune open LLMs from the Llama \cite{grattafiori2024llama3herdmodels} and Qwen \cite{qwen2025qwen25technicalreport} families using LoRA \cite{hu2021loralowrankadaptationlarge}, training separate adapters for each language/domain dataset while keeping the task formulation fixed. 
% We use Llama 3.1 8B Instruct \footnote{\href{https://huggingface.co/meta-llama/Llama-3.1-8B-Instruct}{meta-llama/Llama-3.1-8B-Instruct}} for English (ENG), \textit{Qwen 2.5 7B Instruct}\footnote{\href{https://huggingface.co/Qwen/Qwen2.5-7B-Instruct}{Qwen/Qwen2.5-7B-Instruct}} for Chinese (ZHO), and Qwen 2.5 14B Instruct\footnote{\href{https://huggingface.co/Qwen/Qwen2.5-14B-Instruct}{Qwen/Qwen2.5-14B-Instruct}} for Japanese, Russian, Ukrainian, and Tatar.
We use the instruct versions of Llama 3.1 8B for English, Qwen 2.5 7B for Chinese, and Qwen 2.5 14B for Japanese, Russian, Ukrainian, and Tatar.
The models were chosen based on their support in each respective language. For the lowest resource languages (JPN, RUS, UKR, TAT), a larger model of 14B parameters was chosen to bridge the gap in multilingual support of the LLM. 
% We apply LoRA adapters to the main attention and feed-forward projection layers (\texttt{q\_proj, k\_proj, v\_proj, o\_proj, gate\_proj, up\_proj, down\_proj}), keeping the number of trainable parameters small while retaining a strong adaptation capacity. For efficiency, we load base models in 4-bit precision and train only the LoRA adapters. 
% The training/inference pipeline is shown in Figure \ref{fig:MODEL}.

\section{Results}

\textbf{Experimental setup.}
All experiments used a single NVIDIA A100 GPU. Models were trained on the training set, with results reported on validation and test. Hyperparameters appear in App.~\ref{app:params}.

\textbf{Baselines.} For all subtasks, two baseline models were provided: Kimi K2 Thinking (32B) and Qwen 3 14B. The dataset paper \cite{lee2026dimabsabuildingmultilingualmultidomain} reports Zero-shot and One-shot results with GPT-5 mini and Kimi K2 Thinking, as well as supervised fine-tuning of Qwen 3 14B, Mistral 3 14B, Llama 3.3 70B, and GPT-OSS 120B, which we include for comparison. Test set results are given in App.~\ref{app:bench}.

\subsection{DimASR Results}
% For this subtask, we report the performance on the Test set (Table \ref{tab:test-dimasr}).
Table~\ref{tab:dev-test-dimasr} presents the results for the DimASR subtask in Dev and Test sets. We observe that language-appropriate encoder backbones outperform the provided baselines, except on the Tatar Restaurant dataset, where Kimi K2 Thinking achieves superior performance. Our models outperformed larger LLM-based approaches discussed in the benchmark paper in all English and Chinese domains, along with Japanese Finance, despite having fewer parameters. In Japanese Hotel and in Russian/Ukrainian/Tatar Restaurant, our approach is only outperformed by GPT-OSS (Table~\ref{tab:all_subtasks_bigtable}).

Performance drops are most pronounced in lower-resource languages (notably Tatar), which we attribute to smaller training sets and weaker language-specific pretraining. We report the Pearson Correlation Coefficient (PCC) for deeper analysis. PCC is consistently higher for valence across languages and domains, suggesting that arousal is less directly lexicalized and harder to infer from text. The performance difference between Dev and Test sets can be attributed to split size variations, as shown in App~\ref{app:eda}.

\begin{table}[h!]
\resizebox{0.5\textwidth}{!}{%

    \centering
    \scriptsize
    \setlength{\tabcolsep}{1.4pt} 
    \renewcommand{\arraystretch}{0.95} 

  \begin{tabular}{@{} l l
    S[table-format=1.4] S[table-format=1.4] S[table-format=1.4]
    S[table-format=1.4] S[table-format=1.4] S[table-format=1.4] @{}}
    \toprule
    \textbf{Lang.} & \textbf{Domain} &
    \multicolumn{3}{c}{\textbf{Dev}} & \multicolumn{3}{c}{\textbf{Test}} \\
    \cmidrule(lr){3-5}\cmidrule(lr){6-8}
    & &
    \tiny{\textbf{RMSE$_{VA}$} $\downarrow$} & \tiny{\textbf{PCC$_V$} $\uparrow$} & \tiny{\textbf{PCC$_A$} $\uparrow$} &
    \tiny{\textbf{RMSE$_{VA}$} $\downarrow$} & \tiny{\textbf{PCC$_V$} $\uparrow$} & \tiny{\textbf{PCC$_A$} $\uparrow$} \\
    \midrule

    \multirow{2}{*}{ENG} & Laptop
      & 1.0166 & 0.9174 & 0.5768 & 1.4401 & 0.8520 & 0.5073 \\
    & Restaurant
      & 0.9208 & 0.9295 & 0.6882 & 1.3933 & 0.8757 & 0.6105 \\
    \midrule

    \multirow{3}{*}{ZHO} & Finance
      & 0.5438 & 0.8241 & 0.5981 & 0.5425 & 0.8327 & 0.6436 \\
    & Laptop
      & 0.7443 & 0.8863 & 0.7042 & 0.7457 & 0.8738 & 0.7133 \\
    & Restaurant
      & 0.8007 & 0.8419 & 0.6748 & 1.0023 & 0.8258 & 0.5852 \\
    \midrule

    RUS & Restaurant
      & 1.3808 & 0.8810 & 0.5130 & 1.7236 & 0.7865 & 0.5075 \\
    \midrule

    TAT & Restaurant
      & 1.9234 & 0.5231 & 0.3859 & 2.1144 & 0.5831 & 0.3104 \\
    \midrule

    UKR & Restaurant
      & 1.3692 & 0.9435 & 0.5622 & 1.6724 & 0.8189 & 0.5358 \\
    \midrule

    \multirow{2}{*}{JPN} & Finance
      & 0.9881 & 0.8088 & 0.4398 & 0.9635 & 0.7920 & 0.3497 \\
    & Hotel
      & 0.9844 & 0.9076 & 0.5825 & 0.7484 & 0.9105 & 0.7074 \\
    \bottomrule
  \end{tabular}
}
  \caption{DimASR results per language and domain.}
  \label{tab:dev-test-dimasr}
\end{table}

\subsection{DimASTE Results} 
% For this subtask, we report cF1 on both the Dev and Test sets for completeness (Table \ref{tab:dimaste-dev-test}).

Table~\ref{tab:dimaste-dev-test} presents the results for the DimASTE subtask. All models achieve competitive performance on both sets, outperforming the competition baselines. Compared to the results reported in the benchmark paper (which includes much larger fine-tuned LLMs), our $\leq$14B models display comparable and, in most cases, better results on cF1, except in Tatar Restaurant (outperformed by Llama 3.3 70B) and Japanese Hotel (outperformed by GPT-OSS), as shown in Table~\ref{tab:all_subtasks_bigtable}. The instruct variants of our models seem to help by improving instruction-following and structured output adherence. In addition to fine-tuning, we also report the performance of Qwen 3 14B on the Dev set in a Zero-shot and Few-shot setting. Few-shot worked better than Zero-shot, but the overall performance was underwhelming. The results and insights of this study can be found in App~\ref{app:ablation}.

\begin{table}[h!]
  \centering
  \small
  \setlength{\tabcolsep}{2.5pt}
  \renewcommand{\arraystretch}{1.10}

  \begin{tabular}{@{} l l l S[table-format=1.4] S[table-format=1.4] @{}}
    \toprule
    \textbf{Lang.} & \multicolumn{1}{c}{\textbf{Model}} & \textbf{Domain} &
    \multicolumn{2}{c}{\textbf{cF1 $\uparrow$}} \\
    \cmidrule(lr){4-5}
    & & & \textbf{Dev} & \textbf{Test} \\
    \midrule

    \multirow{2}{*}{ENG} & \multirow{2}{*}{Llama 3.1 8B} & Laptop
      & 0.5962 & 0.5311 \\
    & & Restaurant
      & 0.7668 & 0.6518 \\
    \midrule

    \multirow{2}{*}{ZHO} & \multirow{2}{*}{Qwen 2.5 7B} & Laptop
      & 0.4193 & 0.4646 \\
    & & Restaurant
      & 0.5916 & 0.5042 \\
    \midrule

    JPN & Qwen 2.5 14B & Hotel
      & 0.4879 & 0.5021 \\
    \midrule

    RUS & Qwen 2.5 14B & Restaurant
      & 0.4766 & 0.4988 \\
    \midrule

    UKR & Qwen 2.5 14B & Restaurant
      & 0.4958 & 0.4725 \\
    \midrule

    TAT & Qwen 2.5 14B & Restaurant
      & 0.4272 & 0.3874 \\
    \bottomrule
  \end{tabular}

  \caption{DimASTE results per language and domain.}
  \label{tab:dimaste-dev-test}
\end{table}

The Dev–Test cF1 gaps correspond to variations in review length and structure density (tuples per review), as shown by the Population Stability Index (PSI) heatmap and split statistics in App.~\ref{app:eda}. The effect is significant for Chinese Restaurant, where length and tuples per review vary considerably, and the Test split is denser, increasing the likelihood of missed structures and reducing cF1. Similar patterns appear for English Restaurant and Laptop, where Test reviews are slightly longer or denser than Dev. In lower-resource settings (e.g., Tatar Restaurant), smaller split sizes amplify variance, so moderate length or density differences yield visible Dev–Test fluctuations.

Additionally, we observe a substantial performance gap between English Restaurant and Laptop, despite English being a high-resource language. This can be attributed to the number of NULL entries in the Laptop Train set, which cover nearly half of it (App.~\ref{app:eda}). This introduces a skewed supervision signal that may bias the model toward predicting NULL, even when aspects and/or opinions are explicit.

\subsection{DimASQP Results}
Consistent with the previous subtask, our models show the same behavior when moving to a more complex task. The drop in performance is attributed to the introduction of categories in the extracted fields. We achieve competitive performance, outperforming the provided baselines except in English Laptop (outperformed by Kimi K2 Thinking, as well as GPT-5 mini) and showcasing comparable results against much larger LLMs proposed in the benchmark paper, surpassing their cF1 in most language/domain settings except in Tatar Restaurant (outperformed by Llama 3.3 70B) and Japanese Hotel (outperformed by GPT-OSS) following DimASTE, as shown in Table~\ref{tab:all_subtasks_bigtable}. Zero-shot and Few-shot were also tested using Qwen 2.5 14B on the Dev set, yielding results similar to those of the previous subtask (App~\ref{app:ablation}).

For DimASQP, we additionally fine-tuned larger LLMs than those proposed, namely Llama 3.1 70B, Qwen 2.5 32B, and Qwen 2.5 72B, using the same strategy. They displayed competitive performance on the Dev set across the higher resource languages and fell off in the lower ones. Our proposed efficient models outperformed them consistently. More on this study can be found in App~\ref{app:ablation}.

\begin{table}[h!]
  \centering
  \small
  \setlength{\tabcolsep}{2.5pt}
  \renewcommand{\arraystretch}{1.08}

  \begin{tabular}{@{} l l l S[table-format=1.4] S[table-format=1.4] @{}}
    \toprule
    \textbf{Lang.} & \multicolumn{1}{c}{\textbf{Model}} & \textbf{Domain} &
    \multicolumn{2}{c}{\textbf{cF1 $\uparrow$}} \\
    \cmidrule(lr){4-5}
    & & & \textbf{Dev} & \textbf{Test} \\
    \midrule

    \multirow{2}{*}{ENG} & \multirow{2}{*}{Llama 3.1 8B} & Laptop
      & 0.3556 & 0.2694 \\
    & & Restaurant
      & 0.7560 & 0.5988 \\
    \midrule

    \multirow{2}{*}{ZHO} & \multirow{2}{*}{Qwen 2.5 7B} & Laptop
      & 0.3662 & 0.3703 \\
    & & Restaurant
      & 0.5665 & 0.4544 \\
    \midrule

    JPN & Qwen 2.5 14B & Hotel
      & 0.4218 & 0.3747 \\
    \midrule

    RUS & Qwen 2.5 14B & Restaurant
      & 0.4774 & 0.4369 \\
    \midrule

    UKR & Qwen 2.5 14B & Restaurant
      & 0.4941 & 0.4154 \\
    \midrule

    TAT & Qwen 2.5 14B & Restaurant
      & 0.4033 & 0.3306 \\
    \bottomrule
  \end{tabular}

  \caption{DimASQP results per language and domain.}
  \label{tab:dimasqp-dev-test}
\end{table}

The gap between the Restaurant and Laptop domains becomes larger with the addition of categories. In English and Chinese, models perform much better in the restaurant domain due to higher category/label diversity in the laptop domain compared to other domains. We observe the shift in categories in English Laptop from Train to Dev and Test sets in the PSI heatmap in App.~\ref{app:eda}, which explains the performance drop in this setting. Notably, the Chinese Laptop model performed better without category mentions in the prompt.

\subsection{DimASQP Results via Translation}
Motivated by the performance of our model in the English Restaurant dataset, we translated the Train and Dev sets into English using Qwen 2.5 14B, fine-tuned Llama 3.1 8B on the translated training data, and evaluated it on the translated Dev set. We then mapped the predicted English structures back to the original language using Qwen 2.5 14B. Additionally, we evaluated an English-trained model (trained on English Restaurant) on the translated Dev sets and applied the same back-mapping.

\begin{table}[h!]
\small
  \centering
  \setlength{\tabcolsep}{2.5pt}
  \renewcommand{\arraystretch}{1.08}

  \begin{tabular}{@{} l S[table-format=1.4] S[table-format=1.4] @{}}
    \toprule
    \textbf{Lang.} &
    {\textbf{Translate+Train}} &
    \textbf{{ENG-Model}} \\
    \midrule
    ZHO & 0.36 & 0.26 \\
    \midrule
    RUS & 0.31 & 0.31 \\
    \midrule
    UKR & 0.24 & 0.25 \\
    \midrule
    TAT & 0.20 & 0.22 \\
    \bottomrule
  \end{tabular}

  \caption{Restaurant DimASQP (cF1), Dev set, Llama 3.1 8B: translation training vs. English-trained model.}
  \label{tab:ablation-restaurant-translate-match}
\end{table}

Compared to training on original-language data, translation-based training leads to an overall performance drop, suggesting that translation introduces noise (idiom shift, span drift, or label misalignment). The pre-trained model performed worse only on Chinese Restaurant, consistent with the Chinese–English gap. The Chinese translation gains suggest weaker direct transfer from an English-trained model, so translation-based adaptation can partially compensate despite added noise. In other languages, the pre-trained model achieved better performance, reflecting the difficulty of translating while preserving language-specific idioms. In practice, back-mapping often yields paraphrases rather than original spans, increasing mismatch under exact-match evaluation.

\section{Conclusion}
In our work, we address DimABSA, introduced in SemEval 2026-Task 3. We propose a parameter-efficient methodology that utilizes lightweight transformer encoders for DimASR and LLM backbones with up to 14B parameters for DimASTE and DimASQP. Across languages and domains, our system achieves competitive results on the official evaluation, showing that efficient, smaller-scale models can perform well for DimABSA. In total, we aspire for our methodology to assist future research in the DimABSA setting.

\section{Limitations}
Our system has several limitations. First, we train separate models (or LoRA adapters) for each language/domain dataset, which improves specialization but increases the number of checkpoints to manage and does not directly exploit cross-lingual or cross-domain transfer. Second, structured prediction for DimASTE/DimASQP is evaluated under an exact-match criterion for categorical elements; despite prompt design and greedy decoding, generation can still produce format errors and paraphrased spans, which are heavily penalized by cF1. Third, performance is less stable in low-resource settings where the development sets are small and distribution shifts between splits can be substantial, leading to higher variance and weaker generalization. Finally, our experiments were limited by compute resources (single GPU), restricting broader hyperparameter sweeps and more extensive exploration of larger backbones.

\bibliography{custom}

% \clearpage
\appendix
\section{Data Format}
\label{app:format}

The data format for each subtask is provided in Table \ref{tab:actual-examples-subtasks}.

\begin{table*}[t]
\centering
\small
\begin{tabular}{p{2.2cm} p{12.2cm}}
\hline
\textbf{Subtask} & \textbf{Example (JSON)} \\
\hline
\textbf{DimASR} &
\texttt{\{"ID": "rest26\_aspect\_va\_dev\_1", "Text": "Great diner food and breakfast is served all day", "Aspect\_VA": [\{"Aspect": "diner food", "VA": "7.25\#6.75"\}, \{"Aspect": "breakfast", "VA": "7.25\#6.75"\}]\}} \\
\hline
\textbf{DimASTE} &
\texttt{\{"ID": "rest26\_aste\_dev\_2", "Text": "Customer service was fantastic and food was awesome", "Triplet": [\{"Aspect": "Customer service", "Opinion": "fantastic", "VA": "7.33\#7.33"\}, \{"Aspect": "food", "Opinion": "awesome", "VA": "7.67\#7.67"\}]\}} \\
\hline
\textbf{DimASQP} &
\texttt{\{"ID": "rest26\_asqp\_dev\_1", "Text": "Food and coffee are great", "Quadruplet": [\{"Aspect": "Food", "Category": "FOOD\#QUALITY", "Opinion": "great", "VA": "7.67\#7.83"\}, \{"Aspect": "coffee", "Category": "DRINKS\#QUALITY", "Opinion": "great", "VA": "7.67\#8.00"\}]\}} \\
\hline
\end{tabular}
\caption{Example JSON format for each Subtask.}
\label{tab:actual-examples-subtasks}
\end{table*}

\section{Exploratory Data Analysis}
\label{app:eda}

\begin{figure*}[h]
    \centering
    \includegraphics[width=\textwidth]{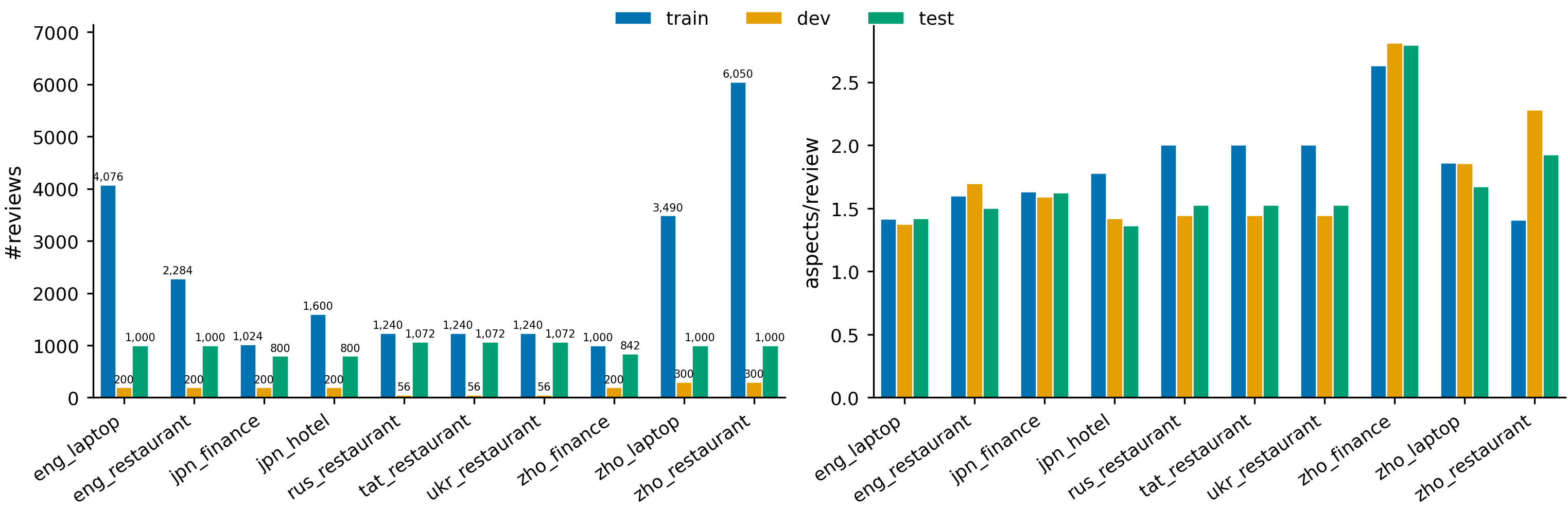}
    \caption{DimASR dataset statistics across splits: number of reviews (left) and aspects per review (right).}
    \label{fig:sub1_size_density}
\end{figure*}

\begin{figure*}[t]
    \centering
    \includegraphics[width=\textwidth]{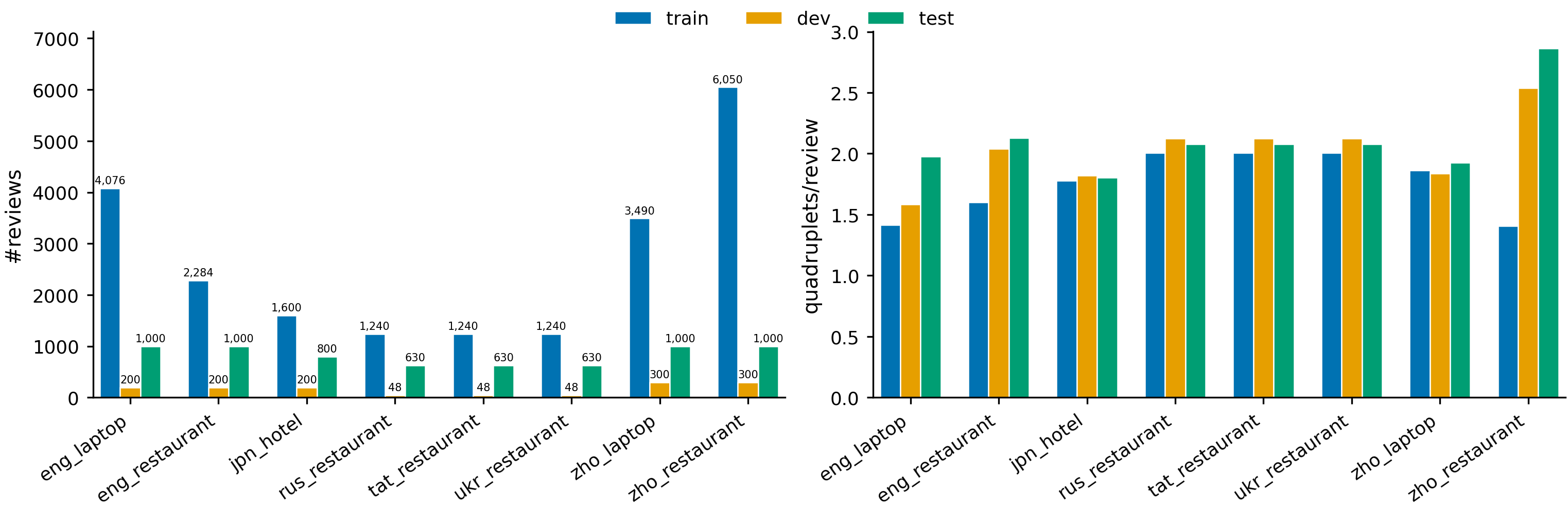}
    \caption{DimASTE and DimASQP dataset statistics across splits: number of reviews (left) and quadruplets per review (right).}
    \label{fig:sub3_size_density}
\end{figure*}

\paragraph{Split Statistics.}
Regarding DimASR, the VA distributions for all the datasets are provided in the dataset paper \cite{lee2026dimabsabuildingmultilingualmultidomain}. Table~\ref{fig:sub1_size_density} summarizes the size of each split, along with the aspects per review for DimASR. We mainly focus our EDA on the other subtasks where the data have more dimensions. The plots presented all contain information about DimASQP, but the datasets are essentially the same as in DimASTE, aside from the keyword change (Triplet/Quadruplet) and the existence of categories. We provide the number of reviews per split and tuples per review in each split in Figure \ref{fig:sub3_size_density}. We can see the substantial difference in size between Dev and Test, which can contribute to performance differences. 

\begin{figure*}[t]
    \centering
    \includegraphics[width=\textwidth]{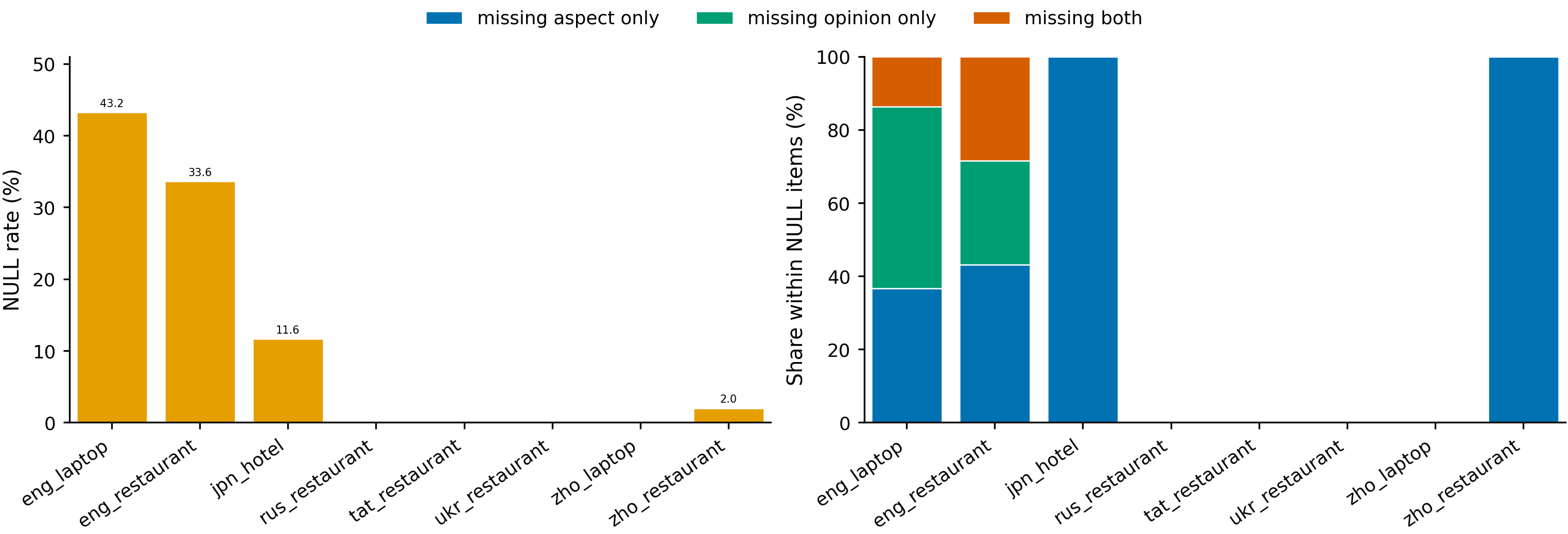}
    \caption{DimASTE and DimASQP NULL analysis on Train set: NULL rate (left) and composition of NULL cases (right).}
    \label{fig:sub3_nulls}
\end{figure*}

\paragraph{NULL Aspect/Opinion instances.}
Following that, we examine the NULL labels in each Train set, as they can bias the model during training (Figure \ref{fig:sub3_nulls}). English datasets contain a large number of NULL labels, with more than half of the NULL instances representing missing opinions. Even though the actual target was missing, sentiment was expressed but not directed at a specific aspect, making these examples valuable as well. The main problem with those NULL entries is the bias they can create for the model to easily output NULL when the aspect/opinion is hard to find in the text. The prompt design was oriented to diminish this bias. 

\paragraph{Splits' distributions shift.}
Lastly, we present a PSI heatmap to identify the differences between each split's distributions (PSI < 0.1: No significant shift, 0.1 - 0.2: Moderate shift, > 0.2: Significant shift). We compare the review lengths, the number of quadruplets per review, and the category distributions between splits for each language and domain combination. This heatmap is useful for explaining specific behaviors in our results.

\section{Benchmark Models}
\label{app:bench}

\paragraph{Competition baselines.} We provide the Hugging Face identifiers for the competition baseline models, which are Kimi K2 Thinking (32B) \footnote{\href{https://huggingface.co/moonshotai/Kimi-K2-Thinking}{moonshotai/Kimi-K2-Thinking}} and Qwen 3 14B \footnote{\href{https://huggingface.co/Qwen/Qwen3-14B}{Qwen/Qwen3-14B}}.

\paragraph{Benchmark paper.} Additionally, we provide the model identifiers for the systems used in the benchmark paper, including Llama 3.3 70B\footnote{\href{https://huggingface.co/meta-llama/Llama-3.3-70B-Instruct}{meta-llama/Llama-3.3-70B-Instruct}}, Mistral 3 14B\footnote{\href{https://huggingface.co/mistralai/Ministral-3-14B-Base-2512}{mistralai/Ministral-3-14B-Base-2512}}, and GPT-OSS 120B\footnote{\href{https://huggingface.co/openai/gpt-oss-120b}{openai/gpt-oss-120b}}. We also report results for GPT-5 mini, accessed via the OpenAI API.

\paragraph{Benchmark Paper Results}
Table~\ref{tab:all_subtasks_bigtable} summarizes the results of the Benchmark paper (including the baseline models).

\begin{table*}[t]
  \centering
  \scriptsize
  \setlength{\tabcolsep}{1.8pt}
  \renewcommand{\arraystretch}{0.92}

  \begin{tabular}{@{} l l l
    S[table-format=1.4] S[table-format=1.4]
    S[table-format=1.4] S[table-format=1.4]
    S[table-format=1.4] S[table-format=1.4] S[table-format=1.4] S[table-format=1.4] @{}}
    \toprule
    \textbf{Subtask} & \textbf{Lang.} & \textbf{Domain} &
    \multicolumn{2}{c}{\textbf{Zero-shot}} &
    \multicolumn{2}{c}{\textbf{One-shot}} &
    \multicolumn{4}{c}{\textbf{Supervised Fine-Tuning}} \\
    \cmidrule(lr){4-5}\cmidrule(lr){6-7}\cmidrule(lr){8-11}
    & & &
    \multicolumn{1}{c}{\textbf{GPT-5 mini}} &
    \multicolumn{1}{c}{\textbf{Kimi K2 Thinking}} &
    \multicolumn{1}{c}{\textbf{GPT-5 mini}} &
    \multicolumn{1}{c}{\textbf{Kimi K2 Thinking}} &
    \multicolumn{1}{c}{\textbf{Qwen 3 14B}} &
    \multicolumn{1}{c}{\textbf{Mistral 3 14B}} &
    \multicolumn{1}{c}{\textbf{Llama 3.3 70B}} &
    \multicolumn{1}{c}{\textbf{GPT-OSS 120B}} \\
    \midrule

    \multirow{10}{*}{DimASR} & ENG & Restaurant & 2.9490 & 2.3432 & 2.3926 & 2.1461 & 2.6427 & 2.6316 & 2.5244 & 1.4605 \\
    & ENG & Laptop     & 3.2115 & 2.6546 & 2.5637 & 2.1893 & 2.8089 & 2.6258 & 2.7354 & 1.5269 \\
    & JPN & Hotel      & 3.1406 & 2.3294 & 2.1607 & 1.7553 & 2.2906 & 2.2999 & 2.6255 & 0.7188 \\
    & JPN & Finance    & 2.6760 & 2.3379 & 1.9243 & 1.6396 & 1.8964 & 2.0700 & 2.4191 & 1.0188 \\
    & RUS & Restaurant & 2.5447 & 2.0630 & 2.0390 & 1.7768 & 2.1528 & 2.3617 & 2.5089 & 1.4775 \\
    & TAT & Restaurant & 2.6645 & 2.3636 & 2.2308 & 1.9380 & 2.6367 & 3.0463 & 2.9165 & 1.7153 \\
    & UKR & Restaurant & 2.5628 & 2.0782 & 2.0438 & 1.7805 & 2.2121 & 2.4592 & 2.5709 & 1.5166 \\
    & ZHO & Restaurant & 2.7125 & 2.2623 & 2.2467 & 1.8959 & 2.0073 & 1.9373 & 2.4463 & 1.0349 \\
    & ZHO & Laptop     & 2.4790 & 2.0426 & 1.9380 & 1.6440 & 1.7706 & 1.8267 & 2.3633 & 0.8032 \\
    & ZHO & Finance    & 2.6547 & 2.9662 & 2.0094 & 1.9652 & 1.4707 & 1.8900 & 2.5632 & 0.6511 \\
    \midrule

    \multirow{8}{*}{DimASTE} & ENG & Restaurant & 0.4993 & 0.5101 & 0.5034 & 0.4920 & 0.4483 & 0.2930 & 0.5418 & 0.5442 \\
    & ENG & Laptop     & 0.4491 & 0.4519 & 0.4874 & 0.4424 & 0.3827 & 0.2736 & 0.4664 & 0.4515 \\
    & JPN & Hotel      & 0.1727 & 0.3148 & 0.2487 & 0.3464 & 0.1622 & 0.1458 & 0.4694 & 0.5397 \\
    & RUS & Restaurant & 0.4016 & 0.4031 & 0.3730 & 0.4242 & 0.3341 & 0.1774 & 0.4590 & 0.4262 \\
    & TAT & Restaurant & 0.3419 & 0.3503 & 0.3159 & 0.3577 & 0.2020 & 0.1154 & 0.4101 & 0.3578 \\
    & UKR & Restaurant & 0.4014 & 0.4082 & 0.3326 & 0.4220 & 0.3099 & 0.1595 & 0.4517 & 0.4250 \\
    & ZHO & Restaurant & 0.3190 & 0.3728 & 0.2425 & 0.3529 & 0.2509 & 0.1446 & 0.4789 & 0.4759 \\
    & ZHO & Laptop     & 0.2372 & 0.2227 & 0.2815 & 0.2494 & 0.2099 & 0.1182 & 0.4344 & 0.4366 \\
    \midrule

    \multirow{8}{*}{DimASQP} & ENG & Restaurant & 0.4036 & 0.3740 & 0.3816 & 0.3746 & 0.2673 & 0.2058 & 0.5048 & 0.5013 \\
    & ENG & Laptop     & 0.2304 & 0.2805 & 0.2842 & 0.2795 & 0.1529 & 0.1293 & 0.2483 & 0.2411 \\
    & JPN & Hotel      & 0.0907 & 0.1309 & 0.1542 & 0.1943 & 0.0400 & 0.0311 & 0.3577 & 0.4151 \\
    & RUS & Restaurant & 0.2508 & 0.2656 & 0.2505 & 0.2963 & 0.1682 & 0.1000 & 0.4118 & 0.3683 \\
    & TAT & Restaurant & 0.1974 & 0.2310 & 0.2025 & 0.2380 & 0.0954 & 0.0611 & 0.3702 & 0.3094 \\
    & UKR & Restaurant & 0.2465 & 0.2974 & 0.2180 & 0.2971 & 0.1641 & 0.0975 & 0.4070 & 0.3663 \\
    & ZHO & Restaurant & 0.2481 & 0.2975 & 0.1891 & 0.2859 & 0.1605 & 0.0934 & 0.4391 & 0.4249 \\
    & ZHO & Laptop     & 0.1356 & 0.1569 & 0.1921 & 0.1900 & 0.1124 & 0.0728 & 0.3506 & 0.3551 \\
    \bottomrule
  \end{tabular}

  \caption{Combined Benchmark paper's results across subtasks. DimASR values are RMSE$_{VA}$; DimASTE and DimASQP values are cF1.}
  \label{tab:all_subtasks_bigtable}
\end{table*}

\section{DimASR Models}
\label{app:reg}
\paragraph{Transformer backbones.}
Table~\ref{tab:dimASR-backbones} summarizes the encoder backbones used for each language, along with their corresponding Hugging Face model identifiers.

\begin{table}[h!]
  \centering
  \scriptsize
  \setlength{\tabcolsep}{6pt}
  \renewcommand{\arraystretch}{1.12}

  \begin{tabular}{@{} l l l @{}}
    \toprule
    \textbf{Language} & \textbf{Model family} & \multicolumn{1}{c}{\textbf{Model ID}} \\
    \midrule
    ENG       & DeBERTa    & \texttt{yangheng/deberta-v3-base-absa-v1.1}\tablefootnote{\href{https://huggingface.co/yangheng/deberta-v3-base-absa-v1.1}{yangheng/deberta-v3-base-absa-v1.1}} \\
    Japanese  & DeBERTa    & \texttt{ku-nlp/deberta-v3-base-japanese}\tablefootnote{\href{https://huggingface.co/ku-nlp/deberta-v3-base-japanese}{ku-nlp/deberta-v3-base-japanese}} \\
    Chinese   & RoBERTa    & \texttt{hfl/chinese-roberta-wwm-ext}\tablefootnote{\href{https://huggingface.co/hfl/chinese-roberta-wwm-ext}{hfl/chinese-roberta-wwm-ext}} \\
    Russian   & BERT-style & \texttt{DeepPavlov/rubert-base-cased}\tablefootnote{\href{https://huggingface.co/DeepPavlov/rubert-base-cased}{DeepPavlov/rubert-base-cased}} \\
    Ukrainian & XLM-R      & \texttt{FacebookAI/xlm-roberta-base}\tablefootnote{\href{https://huggingface.co/FacebookAI/xlm-roberta-base}{FacebookAI/xlm-roberta-base}} \\
    Tatar     & XLM-R      & \texttt{FacebookAI/xlm-roberta-base}\tablefootnote{\href{https://huggingface.co/FacebookAI/xlm-roberta-base}{FacebookAI/xlm-roberta-base}} \\
    \bottomrule
  \end{tabular}

  \caption{Language-specific encoder backbones used for DimASR.}
  \label{tab:dimASR-backbones}
\end{table}

\paragraph{Regression head.}
Given the aspect-conditioned input, the encoder backbone produces contextual token representations $\{\mathbf{h}_t\}_{t=1}^{T}$, where $t$ indexes tokens and $\mathbf{h}_t$ is the representation of token $t$. We compute an attention-pooled sequence representation $\mathbf{z}$ using a trainable scoring vector $\mathbf{w}$:
\begin{align}
s_t &= \mathbf{w}^{\top}\mathbf{h}_t,\\
\alpha_t &= \frac{\exp(s_t)}{\sum_{j=1}^{T}\exp(s_j)},\\
\mathbf{z} &= \sum_{t=1}^{T}\alpha_t\mathbf{h}_t,
\end{align}
where $s_t$ is the attention score and $\alpha_t$ is the corresponding normalized weight (padding tokens are masked before normalization). We then predict valence and arousal using two linear regression heads:
\begin{align}
\tilde{v} &= \mathbf{W}_v\mathbf{z}+b_v,\\
\tilde{a} &= \mathbf{W}_a\mathbf{z}+b_a,
\end{align}
where $\mathbf{W}_v,\mathbf{W}_a$ and $b_v,b_a$ are trainable parameters, and $(\tilde{v},\tilde{a})$ are the model outputs for valence and arousal.

\paragraph{Training objective.}
Let $\mathbf{y}_i=(v_i,a_i)$ denote the gold VA pair and $\tilde{\mathbf{y}}_i=(\tilde{v}_i,\tilde{a}_i)$ the corresponding model output for instance $i$ in a minibatch of size $B$. During training, VA labels are linearly normalized to $[0, 1]$; at inference time, we rescale and clip predictions to enforce valid ranges. We optimize a weighted function of Mean Squared Error (MSE) and Concordance Correlation Coefficient (CCC), and add a VA-guided triplet regularizer using the standard hinge triplet objective.

\begin{align}
\mathcal{L}_{\mathrm{MSE}} &= \frac{1}{B}\sum_{i=1}^{B}\left\lVert \tilde{\mathbf{y}}_i-\mathbf{y}_i \right\rVert_2^2 .
\end{align}
For each dimension $k\in\{v,a\}$, let $\tilde{\mathbf{y}}^{(k)}$ and $\mathbf{y}^{(k)}$ be the vectors of predictions and gold labels over the minibatch. CCC is:
\begin{align}
\mathrm{CCC}_k &=
\frac{2\sigma_{\tilde{y}^{(k)}y^{(k)}}}
{\sigma^2_{\tilde{y}^{(k)}}+\sigma^2_{y^{(k)}}+\left(\mu_{\tilde{y}^{(k)}}-\mu_{y^{(k)}}\right)^2},
\end{align}
where $\mu$ denotes the minibatch mean, $\sigma^2$ the minibatch variance, and $\sigma_{\tilde{y}^{(k)}y^{(k)}}$ the minibatch covariance. We aggregate the two dimensions as $\mathrm{CCC}=\lambda_v\mathrm{CCC}_v+\lambda_a\mathrm{CCC}_a$, where $\lambda_v,\lambda_a$ weight the two dimensions.

To regularize the representation space, we sample triplets $(i,p,n)$ within a minibatch based on distances in the VA label space of the gold pairs (positives close to the anchor, negatives far) and apply:

\begin{align}
\ell_{i,p,n} &= \max\!\Big(0,\,
\lVert \mathbf{z}_i-\mathbf{z}_p\rVert_2
- \lVert \mathbf{z}_i-\mathbf{z}_n\rVert_2
+ m\Big), \\
\mathcal{L}_{\mathrm{tri}} &=
\frac{1}{|\mathcal{T}|}\sum_{(i,p,n)\in\mathcal{T}} \ell_{i,p,n},
\end{align}
where $\mathcal{T}$ denotes the set of sampled triplets and $m>0$ is the margin. The final objective is:
\begin{align}
\mathcal{L}
&= (1-\beta)\,\mathcal{L}_{\mathrm{base}} + \beta\,\mathcal{L}_{\mathrm{tri}}, \\
\mathcal{L}_{\mathrm{base}}
&= \gamma\,\mathcal{L}_{\mathrm{MSE}} + (1-\gamma)\big(1-\mathrm{CCC}\big),
\end{align}
where $\gamma\in[0,1]$ controls the trade-off between MSE and CCC in the base regression loss, and $\beta\in[0,1]$ balances the base loss and the contrastive VA regularizer.

\section{DimASTE and DimASQP Models}
\label{app:gen}
We use Llama 3.1 8B Instruct \footnote{\href{https://huggingface.co/meta-llama/Llama-3.1-8B-Instruct}{meta-llama/Llama-3.1-8B-Instruct}} for English, Qwen 2.5 7B Instruct \footnote{\href{https://huggingface.co/Qwen/Qwen2.5-7B-Instruct}{Qwen/Qwen2.5-7B-Instruct}} for Chinese, and Qwen 2.5 14B Instruct \footnote{\href{https://huggingface.co/Qwen/Qwen2.5-14B-Instruct}{Qwen/Qwen2.5-14B-Instruct}} for Japanese, Russian, Ukrainian, and Tatar.

\section{Language Specific Prompts}
\label{app:prompts}
We construct prompts using the native chat template of each backbone to improve instruction-following and JSON adherence. For Llama-style models, we use the \texttt{<|begin\_of\_text|>} / \texttt{<|eot\_id|>} turn delimiters, while for Qwen-style models, we use \texttt{<|im\_start|>} / \texttt{<|im\_end|>}. The full training prompts for DimASQP can be found in Table \ref{tab:prompts-by-lang-domain}. The DimASTE training prompts are the same, without the category mentions. For inference, we use exactly the same prompt without the answer.

\section{Experimental Settings}
\label{app:params}

\subsection{DimASR Settings.}

In all experiments, we set $\gamma=0.3$ in $\mathcal{L}_{\mathrm{base}}$, and weight the CCC aggregation as
$\lambda_v=0.3$ and $\lambda_a=0.7$.
For the final objective, we set $\beta=0.05$, yielding a $0.95/0.05$ weighting between
$\mathcal{L}_{\mathrm{base}}$ and $\mathcal{L}_{\mathrm{tri}}$.
The triplet term is computed on detached embeddings (no gradient flow through $\mathbf{z}$ for $\mathcal{L}_{\mathrm{tri}}$).

The hyperparameters used for training can be found in Table \ref{tab:hyperparamsdimASR}. We use a two-stage learning-rate schedule: a step-wise linear warmup for the first $10\%$ of training steps (updated every mini-batch), followed by an epoch-wise ReduceLROnPlateau schedule applied after each validation epoch, monitoring $\mathrm{RMSE}_{\mathrm{VA}}$ (factor $0.5$, patience $2$).

\begin{table}[h]
\centering
\small
\begin{tabular}{ll}
\hline
\textbf{Hyperparameter} & \textbf{Value} \\
\hline
Learning rate ($\eta$) & $2e-5$ \\
Batch size & 16 \\
Dropout & 0.3 \\
Warmup ratio & 0.1 \\
Early stopping patience & 5 \\
Max sequence length & 128 \\
Optimizer & AdamW \\
Scheduler & Warmup + ReduceLROnPlateau \\
Epochs & 30 \\
\hline
\end{tabular}
\caption{Training hyperparameters for the DimASR models.}
\label{tab:hyperparamsdimASR}
\end{table}

\subsection{DimASTE \& DimASQP Settings}
We apply LoRA adapters to the main attention and feed-forward projection layers, keeping the number of trainable parameters small while retaining a strong adaptation capacity. For efficiency, we load base models in 4-bit precision and train only the LoRA adapters. The hyperparameters used for instruction fine-tuning our LLMs can be found in Table \ref{tab:llm_hparams}.

\begin{table}[h]
\centering
\small
\begin{tabular}{ll}
\hline
\textbf{Hyperparameter} & \textbf{Value} \\
\hline
Train epochs ($E$) & 1 \\
Per-device batch size & 2 \\
Gradient accumulation steps & 4 \\
Effective batch (per device) & $2 \times 4 = 8$ \\
Learning rate ($\eta$) & $2e-4$ \\
Weight decay & $1 \times 10^{-4}$ \\
Warmup ratio & 0.03 \\
LR scheduler & linear \\
Optimizer & paged\_adamw\_32bit \\
Max sequence length & 2048 \\
Quantization & 4-bit \\
Precision & bf16 \\
Max grad norm & 0.3 \\
\hline
LoRA rank ($r$) & 16 \\
LoRA $\alpha$ & 32 \\
LoRA dropout & 0.2 \\
LoRA target modules &
\begin{tabular}[c]{@{}l@{}}
{\scriptsize\texttt{[q\_proj, k\_proj, v\_proj,}}\\
{\scriptsize\texttt{o\_proj, gate\_proj, up\_proj,}}\\
{\scriptsize\texttt{down\_proj]}}
\end{tabular} \\
\hline
\end{tabular}
\caption{LLM fine-tuning configuration (4-bit loading + LoRA adapters) for DimASTE \& DimASQP.}
\label{tab:llm_hparams}
\end{table}

\section{Additional Ablation Studies}
\label{app:ablation}

\subsection{Zero-shot \& Few-shot}
We also tested our methodology in a zero-shot/few-shot scenario for DimASTE and DimASQP using Qwen 2.5 14B Instruct on the Dev set (Tables \ref{tab:subtask2-zeroshot-fewshot-dev} \& \ref{tab:subtask3-zeroshot-fewshot-dev}). For few-shot prompting, we randomly sample three demonstrations from the training set of the corresponding language-domain pair and prepend them under a single instruction, following the same output schema as the task. We include each demonstration as an input-output pair (review plus its gold JSON) and apply the same post-processing as in our main pipeline before scoring. We decode greedily to improve format stability and JSON adherence. 

Few-shot performed much better than zero-shot across all languages and domains, but it still couldn't reach the performance of fine-tuning. These techniques yield better results in high-resource languages such as English and Chinese. Notably, for English Laptop, few-shot prompting approaches the fine-tuned Dev performance, suggesting that in some cases, in-context demonstrations can leverage the model’s prior domain knowledge and reduce the need for full fine-tuning; however, this trend is less consistent in lower-resource languages.

\begin{table}[h!]
  \centering
  \small
  \setlength{\tabcolsep}{6pt}
  \renewcommand{\arraystretch}{1.12}

  \begin{tabular}{@{} l l S[table-format=1.2] S[table-format=1.2] @{}}
    \toprule
    \textbf{Lang.} & \textbf{Domain} &
    \multicolumn{2}{c}{\textbf{Dev cF1} $\uparrow$} \\
    \cmidrule(lr){3-4}
    & & {Zero-shot} & {Few-shot} \\
    \midrule

    \multirow{2}{*}{ENG} & Laptop     & 0.48 & 0.59 \\
                         & Restaurant & 0.53 & 0.59 \\
    \midrule

    \multirow{2}{*}{ZHO} & Laptop     & 0.23 & 0.27 \\
                         & Restaurant & 0.37 & 0.43 \\
    \midrule

    JPN & Hotel      & 0.17 & 0.19 \\
    \midrule
    RUS & Restaurant & 0.33 & 0.37 \\
    \midrule
    UKR & Restaurant & 0.33 & 0.35 \\
    \midrule
    TAT & Restaurant & 0.15 & 0.16 \\
    \bottomrule
  \end{tabular}

  \caption{DimASTE Dev set results in zero-shot and few-shot prompting (cF1) using Qwen 2.5 14B Instruct.}
  \label{tab:subtask2-zeroshot-fewshot-dev}
\end{table}

\begin{table}[h!]
  \centering
  \small
  \setlength{\tabcolsep}{6pt}
  \renewcommand{\arraystretch}{1.12}

  \begin{tabular}{@{} l l S[table-format=1.2] S[table-format=1.2] @{}}
    \toprule
    \textbf{Lang.} & \textbf{Domain} &
    \multicolumn{2}{c}{\textbf{Dev cF1} $\uparrow$} \\
    \cmidrule(lr){3-4}
    & & {Zero-shot} & {Few-shot} \\
    \midrule

    \multirow{2}{*}{ENG} & Laptop     & 0.22 & 0.32 \\
                         & Restaurant & 0.42 & 0.51 \\
    \midrule

    \multirow{2}{*}{ZHO} & Laptop     & 0.14 & 0.18 \\
                         & Restaurant & 0.23 & 0.31 \\
    \midrule

    JPN & Hotel      & 0.06 & 0.11 \\
    \midrule
    RUS & Restaurant & 0.25 & 0.30 \\
    \midrule
    UKR & Restaurant & 0.24 & 0.25 \\
    \midrule
    TAT & Restaurant & 0.07 & 0.15 \\
    \bottomrule
  \end{tabular}

  \caption{DimASQP Dev set results in zero-shot and few-shot prompting (cF1) using Qwen 2.5 14B Instruct.}
  \label{tab:subtask3-zeroshot-fewshot-dev}
\end{table}    

\subsection{Fine-tuning larger LLMs}
Using the same strategy and hyperparameters, we tried to utilize larger LLMs for DimASQP, such as Llama 3.1 70B Instruct \footnote{\href{https://huggingface.co/meta-llama/Llama-3.1-70B-Instruct}{meta-llama/Llama-3.1-70B-Instruct}}, Qwen 2.5 32B Instruct\footnote{\href{https://huggingface.co/Qwen/Qwen2.5-32B-Instruct}{Qwen/Qwen2.5-32B-Instruct}}, and Qwen 2.5 32B Instruct\footnote{\href{https://huggingface.co/Qwen/Qwen2.5-72B-Instruct}{Qwen/Qwen2.5-72B-Instruct}}. Due to resource constraints, we couldn't dive further to explore different hyperparameters and make multiple runs.

\begin{table}[!]
  \centering
  \small
  \setlength{\tabcolsep}{4pt}
  \renewcommand{\arraystretch}{1.12}

  \begin{tabular}{@{} l l l S[table-format=1.4] @{}}
    \toprule
    \textbf{Lang.} & \multicolumn{1}{c}{\textbf{Model}} & \textbf{Domain} & {\textbf{Dev cF1} $\uparrow$} \\
    \midrule

    \multirow{2}{*}{ENG} &
    \multirow{2}{*}{Llama 3.1 70B} &
    Laptop     & 0.3373 \\
    & & Restaurant & 0.7441 \\
    \midrule

    \multirow{4}{*}{ZHO} &
    \multirow{2}{*}{Qwen 2.5 32B} &
    Laptop     & 0.3119 \\
    & & Restaurant & 0.5220 \\
    & \multirow{2}{*}{Qwen 2.5 72B} &
    Laptop     & 0.3675 \\
    & & Restaurant & 0.5508 \\
    \midrule

    \multirow{2}{*}{JPN} & Qwen 2.5 32B & \multirow{2}{*}{Hotel}
      & 0.3044 \\
    & Qwen 2.5 72B &
      & 0.3414 \\
    \midrule

    \multirow{2}{*}{RUS} & Qwen 2.5 32B & \multirow{2}{*}{Restaurant}
      & 0.3714 \\
    & Qwen 2.5 72B &
      & 0.4332 \\
    \midrule

    \multirow{2}{*}{UKR} & Qwen 2.5 32B & \multirow{2}{*}{Restaurant}
      & 0.3610 \\
    & Qwen 2.5 72B &
      & 0.4328 \\
    \midrule

    \multirow{2}{*}{TAT} & Qwen 2.5 32B & \multirow{2}{*}{Restaurant}
      & 0.2812 \\
    & Qwen 2.5 72B &
      & 0.3467 \\
    \bottomrule
  \end{tabular}

  \caption{DimASQP development set results (cF1) after lora-tuning larger LLMs.}
  \label{tab:subtask3-llama70b-qwen32b-qwen72b-dev}
\end{table}

Under the same fine-tuning recipe, larger backbones did not consistently improve cF1 over our $\leq$14B models, and in several lower-resource settings, they performed worse. This suggests that scaling benefits may require more careful hyperparameter tuning and/or longer training for larger models.

\begin{figure*}[t]
    \centering
    \includegraphics[width=\textwidth]{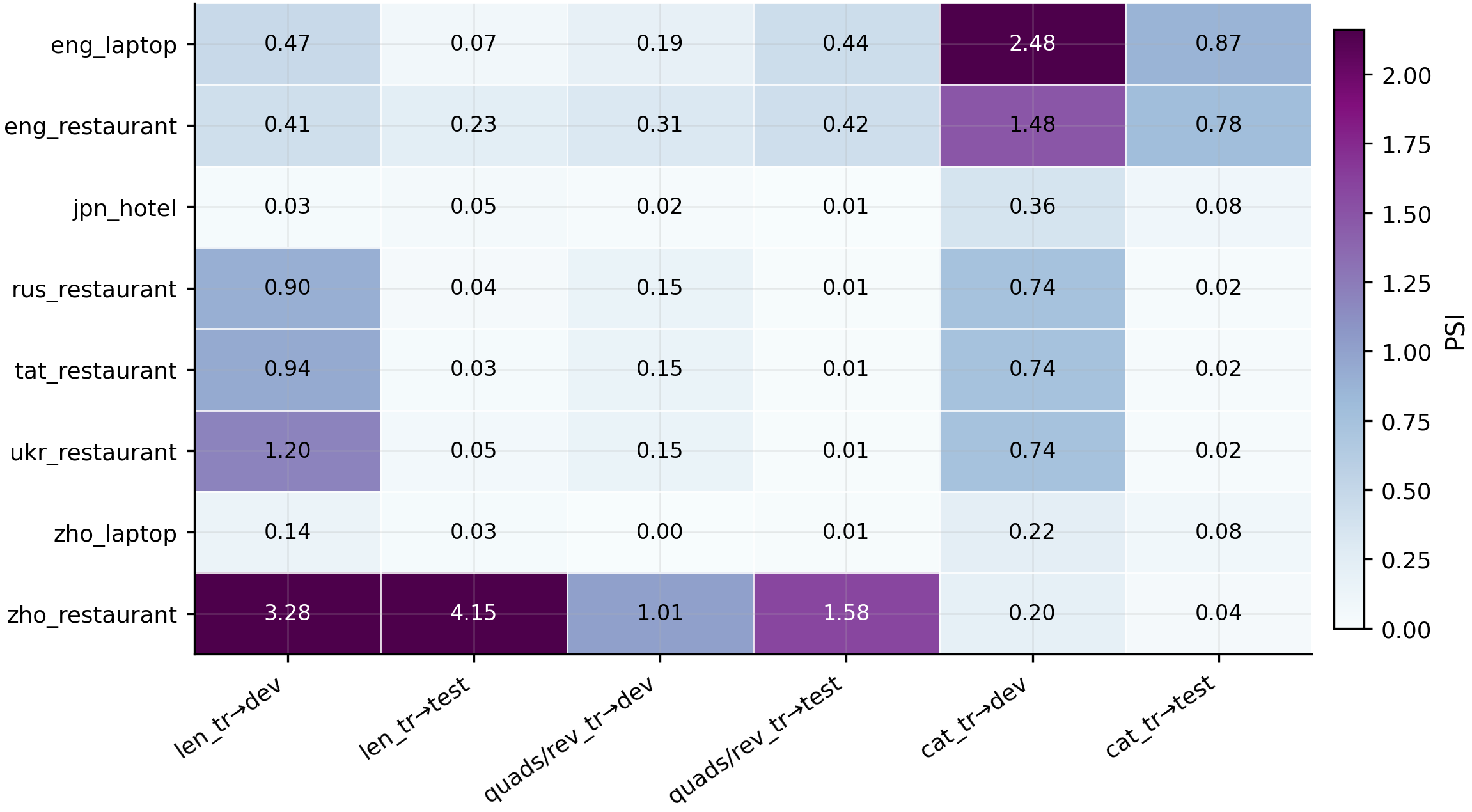}
    \caption{DimASTE and DimASQP distribution shift measured by PSI between Train and Dev/Test across features (review length, quadruplets/review, and category).}
    \label{fig:sub3_psi}
\end{figure*}

\clearpage

% Create a real table number in THIS document, so \ref works
\refstepcounter{table}
\label{tab:prompts-by-lang-domain}

% Optional: add an entry to the List of Tables
\addcontentsline{lot}{table}{\protect\numberline{\thetable}Training prompts per language/domain and full chat templates}

% Now include the PDF pages (these pages will show the MAIN doc page numbers)



\section*{Supplementary material}

# Prompt Templates Used for Training (DimASQP)

| Lang_Domain | Instruction | Training chat template (user + assistant) |
|---|---|---|
| **eng_restaurant** | ```
You are an expert sentiment analysis tool. Your task is to extract all
↪ quadruplets from the given review.
Here are the rules:
1. You MUST format your answer as a single, valid JSON list of objects.
2. Each object MUST have the keys: 'Aspect', 'Category', 'Opinion',
↪ 'Valence', and 'Arousal'.
3. 'Aspect' and 'Opinion' MUST be exact substrings from the input text.
   - Copy the text EXACTLY as it appears, including spacing,
↪   punctuation errors, or typos.
   - Do NOT correct or normalize the text.
3b. Always try your best to find explicit 'Aspect' and 'Opinion'
↪ mentions.
   - The 'Opinion' may appear before or after the 'Aspect' in the
↪   text; always extract the full 'Aspect' and 'Opinion',
↪   regardless of order in the sentence.
   - Use 'NULL' only if a real 'Aspect' or 'Opinion' span truly does
↪   not exist.
3c. 'Opinion' MUST be complete and include all adjectives, negations,
↪ and modifiers.
4. 'Category' MUST be one of the following values: [{CATEGORIES}].
5. 'Valence' and 'Arousal' MUST be numbers from 1.00 to 9.00, with up
↪ to two decimal places.
6. Your entire response MUST be *only* the JSON list.
Now, perform the task for the following review:
``` | ```
<|begin_of_text|><|start_header_id|>user<|end_header_id|>
{instruction}
Review: "{text}"<|eot_id|>
<|start_header_id|>assistant<|end_header_id|>
{answer_json}<|eot_id|>
``` |
| **eng_laptop** | ```
You are an expert sentiment analysis tool. Your task is to extract all
↪ quadruplets from the given review.
Here are the rules:
1. You MUST format your answer as a single, valid JSON list of objects.
2. Each object MUST have the keys: 'Aspect', 'Category', 'Opinion',
↪ 'Valence', and 'Arousal'.
3. 'Aspect' and 'Opinion' MUST be exact substrings from the input text.
   - Copy the text EXACTLY as it appears, including spacing,
↪   punctuation errors, or typos.
   - Do NOT correct or normalize the text.
3b. Always try your best to find explicit 'Aspect' and 'Opinion'
↪ mentions.
   - The 'Opinion' may appear before or after the 'Aspect' in the
↪   text; always extract the full 'Aspect' and 'Opinion',
↪   regardless of order in the sentence.
   - Use 'NULL' only if a real 'Aspect' or 'Opinion' span truly does
↪   not exist.
3c. 'Opinion' MUST be complete and include all adjectives, negations,
↪ and modifiers.
3d. 'Aspect' MUST be complete and include descriptive adjectives.
4. 'Category' MUST be one of the following values: [{CATEGORIES}].
5. 'Valence' and 'Arousal' MUST be numbers from 1.00 to 9.00, with up
↪ to two decimal places.
6. Your entire response MUST be *only* the JSON list.
Now, perform the task for the following review:
``` | ```
<|begin_of_text|><|start_header_id|>user<|end_header_id|>
{instruction}
Review: "{text}"<|eot_id|>
<|start_header_id|>assistant<|end_header_id|>
{answer_json}<|eot_id|>
``` |
| **zho_restaurant** | ```
你是一個精準且嚴格的中文情感四元組抽取模型。
你的任務是從評論中抽取所有可能且明確存在的四元組,
↪ 且不要遺漏任何明確的 Aspect-Opinion 組合, 同時嚴格避免假陽性。

每個四元組包含以下欄位 (欄位順序必須固定):
"Aspect" (評論中出現的詞, 必須完全照抄原文字串)
"Category" (必須與 {CATEGORIES} 中的某一項完全一致, 不可創造、
↪ 改寫或合併新分類)
"Opinion" (完整且連續描述該 Aspect 的情緒片段, 必須完全照抄原文字串,
↪ 包括所有否定、程度、副詞或連續修飾語)
"Valence" (情緒正負度, 1.00–9.00, 小數兩位,
↪ 基於語義度的主觀連續估計)
"Arousal" (情緒喚起度, 1.00–9.00, 小數兩位,
↪ 基於語義度的主觀連續估計)

請遵守以下規則:
1. 輸出必須是一個 JSON 陣列, 不得包含任何非 JSON 文字
2. Aspect 與 Opinion 必須完全照抄原文, 不可修改、標準化、裁切或推測
3. 若評論中存在多個不同的 Aspect 或 Opinion, 需分別生成多個 JSON 物件;
↪ 若同一 Aspect 對多個不同 Opinion, 也需分別生成多個四元組
4. 僅當評論中存在明確情緒表達, 但無法在原文中找到任何可對應的 Aspect 或
↪ Opinion 文字片段時, 才允許填寫 "NULL"
5. 若評論中不存在任何明確可抽取的情感四元組, 輸出空陣列 []
6. 嚴格避免假陽性: 僅在 Aspect 與 Opinion
↪ 均在原文中明確存在時才生成四元組
7. 請檢查評論中所有可能的 Aspect-Opinion 組合,
↪ 務必完整覆蓋所有明確表達的情感
8. 僅輸出 JSON, 不要附加任何說明、分析或解釋

請根據以下評論抽取情感四元組:
``` | ```
<|im_start|>user
{instruction}
評論: "{text}"
<|im_end|>
<|im_start|>assistant
{answer_json}
<|im_end|>
``` |

| Lang/Domain | Instruction (original prompt) | Training chat template (user + assistant) |
|---|---|---|
| **zho_laptop** | 你是一個精準且嚴格的中文情感四元組抽取模型。<br>你的任務是從評論中抽取所有可能且明確存在的四元組，且不要遺漏任何明確的 Aspect-Opinion 組合，同時嚴格避免假陽性。<br><br>每個四元組包含以下欄位（欄位順序必須固定）：<br>"Aspect"（評論中出現的詞，必須完全照抄原文字串）<br>"Category"<br>"Opinion"（完整且連續描述該 Aspect 的情緒片段，必須完全照抄原文字串，包括所有否定、程度、副詞或連續修飾語）<br>"Valence"（情緒正負屬度，1.00~9.00，小數兩位，基於語義屬度的主觀連續估計）<br>"Arousal"（情緒屬起屬度，1.00~9.00，小數兩位，基於語義屬度的主觀連續估計）<br><br>請遵守以下規則：<br>1. 輸出必須是一個 JSON 陣列，不得包含任何非 JSON 文字<br>2. Aspect 與 Opinion 必須完全照抄原文，不可修改、標準化、裁切或推測<br>3. 若評論中存在多個不同的 Aspect 或 Opinion，需分屬生成多個 JSON 物件；若同一 Aspect 對應多個不同 Opinion，也需分屬生成多個四元組<br>4. 僅當評論中存在明確情緒表達，但無法在原文中找到任何可對應的 Aspect 或 Opinion 文字片段時，才允許填寫 "NULL"<br>5. 若評論中不存在任何明確可抽取的情感四元組，輸出空陣列 []<br>6. 嚴格避免假陽性：僅在 Aspect 與 Opinion 均在原文中明確存在時才生成四元組<br>7. 請檢查評論中所有可能的 Aspect-Opinion 組合，務必完整覆蓋所有明確表達的情感<br>8. 僅輸出 JSON，不要附加任何屬明、分析或解釋<br><br>請根據以下評論抽取情感四元組： | `<\|im_start\|>user`<br>`{instruction}`<br><br>`評論: "{text}"`<br>`<\|im_end\|>`<br>`<\|im_start\|>assistant`<br>`{answer_json}`<br>`<\|im_end\|>` |
| **jpn_hotel** | あなたは日本語レビューから感情四つ組を抽出するモデルです。以下の規則に屬密に屬い、入力テキストから四つ組をすべて抽出してください。【使用可能なカテゴリー屬】{CATEGORIES}【抽出項目】1. "Aspect": テキスト内に屬際に出現する屬象語（原文を一字一句そのまま抽出）2. "Category": 上記カテゴリー屬から最も適切なカテゴリ名を選ぶ 3. "Opinion": Aspect に屬する評屬表現（原文をそのまま抽出し、否定・程度語も含める）4. "Valence": 感情の正負屬度（1.00屬9.00）5. "Arousal": 感情の屬醒度（1.00屬9.00）【出力規則】- JSON 配列のみを出力し、余計な文字を一切含めないこと。- 原文に存在しない語句を作らないこと。- 明確な Aspect または Opinion がない場合は "NULL" を使用する。- 抽出すべき四つ組が存在しない場合は [] を返す。- 屬数ある場合は要素を分けてすべて出力する。 | `<\|im_start\|>user`<br>`{instruction}`<br><br>`レビュー本文: "{text}"`<br>`<\|im_end\|>`<br>`<\|im_start\|>assistant`<br>`{answer_json}`<br>`<\|im_end\|>` |
| **rus_restaurant** | Вы — эксперт по анализу настроений. Ваша задача — извлечь все эмоциональные четверки из данного пользовательского отзыва. Вот правила:<br>1. Вы ДОЛЖНЫ форматировать ответ как один корректный JSON-список объектов.<br>2. Каждый объект ДОЛЖЕН содержать ключи: 'Aspect', 'Category', 'Opinion', 'Valence' и 'Arousal'.<br>3. 'Aspect' и 'Opinion' ДОЛЖНЫ быть точными подстроками из исходного текста.<br>  - Копируйте текст ТОЧНО как он есть, включая пробелы, пунктуацию и ошибки.<br>  - НЕ исправляйте и не нормализуйте текст.<br>  - 'Opinion' может идти перед или после 'Aspect'; извлекайте полные 'Aspect' и 'Opinion', независимо от порядка в предложении.<br>  - 'Opinion' ДОЛЖНО быть полным, включая все прилагательные, отрицания и модификаторы.<br>4. 'Category' ДОЛЖЕН быть одним из следующих значений: [{CATEGORIES}].<br>5. 'Valence' и 'Arousal' ДОЛЖНЫ быть числами от 1.00 до 9.00 с точностью до двух знаков после запятой.<br>6. Ваш ответ должен содержать *только* JSON-список.<br>Теперь выполните задачу для следующего отзыва: | `<\|im_start\|>user`<br>`{instruction}`<br><br>`Отзыв: "{text}"`<br>`<\|im_end\|>`<br>`<\|im_start\|>assistant`<br>`{answer_json}`<br>`<\|im_end\|>` |

| Lang/Domain | Instruction (original prompt) | Training chat template (user + assistant) |
|---|---|---|
| **ukr_restaurant** | Ти є високоточним інструментом аспектно-орієнтованого аналізу тональності.<br>Твоє завдання — витягнути всі релевантні чотирики (Aspect–Category–Opinion–Valence–Arousal) з поданого тексту відгуку.<br><br>Правила:<br>1. Відповідь має бути єдиним валідним JSON-масивом об'єктів.<br>2. Кожен об'єкт відповідає одній парі Aspect–Opinion, що виражає одну окрему оцінку.<br>3. Кожен об'єкт повинен містити рівно такі ключі: "Aspect", "Category", "Opinion", "Valence", "Arousal".<br>4. "Aspect" і "Opinion" повинні бути точними підрядками з тексту відгуку — копіюй символ у символ, без змін, нормалізації чи перефразування.<br>5. "Opinion" має включати всі прикметники, заперечення та підсилювачі, що формують емоційну оцінку аспекту.<br>6. "Category" повинна бути рівно одним значенням зі списку дозволених категорій: [{CATEGORIES}]. Жодні інші значення заборонені.<br>7. "Valence" і "Arousal" — дійсні числа від 1.00 до 9.00 включно, обов'язково з двома десятковими знаками. Valence: 1.00 — вкрай негативна, 9.00 — вкрай позитивна оцінка. Arousal: 1.00 — низька емоційна інтенсивність, 9.00 — дуже висока інтенсивність.<br>8. Якщо у відгуку немає жодної релевантної пари Aspect–Opinion, поверни порожній JSON-масив [].<br>9. Відповідь повинна містити лише JSON-масив — без пояснень, коментарів або будь-якого додаткового тексту.<br><br>Ось текст відгуку: | `<\|im_start\|>user`<br>`{instruction}`<br><br>`Текст відгуку: "{text}"`<br>`<\|im_end\|>`<br>`<\|im_start\|>assistant`<br>`{answer_json}`<br>`<\|im_end\|>` |
| **tat_restaurant** | Син — татар телендәге күзәтү текстларыннан аспект-дүртлекләрне чыгару өчен махсуслашкан система.<br>Түбәндәге кагыйдәләрне төгәл үтә:<br>1) Җавап БАРЫ тик дөрес JSON-массив булырга тиеш, башка бернинди текст язма.<br>2) һәр элементта мәҗбүри кырлар: "Aspect","Category","Opinion","Valence","Arousal".<br>3) "Aspect" һәм "Opinion" тексттан үзгәртүсез күчерелә.<br>4) "Aspect" һәм "Opinion" теләсә нинди тәртиптә очрый ала, ләкин текстта мөмкин булган һӘР аспект–фикер парын аерым дүртлек итеп чыгару мәҗбүри.<br>5) "Opinion" барлык сыйфатларны, көчәйткечләрне һәм кире формаларны тулысынча эченә ала.<br>6) "Category" түбәндәгеләрнең берсе генә булырга тиеш: [{CATEGORIES}].<br>7) "Valence" һәм "Arousal" 1.00–9.00 диапазонында, ике дөңгәләк санга кадәр күрсәтелә.<br><br>Күзәтү тексты: | `<\|im_start\|>user`<br>`{instruction}`<br><br>`Текст бәяләмәсе "{text}"`<br>`<\|im_end\|>`<br>`<\|im_start\|>assistant`<br>`{answer_json}`<br>`<\|im_end\|>` |



\end{document}